# Color Image Enhancement Method Based on Weighted Image Guided Filtering

MU Qi[1,2], WEI Yanyan[1], LI Zhanli[1]

(1. College of computer science and technology, Xi'an University of Science and Technology, Xi'an 710054; 2. School of Mechanical Engineering, Xi'an University of Science and Technology, Xi'an 710054)

**Abstract:** A novel color image enhancement method is proposed based on Retinex to enhance color images under non-uniform illumination or poor visibility conditions. Different from the conventional Retinex algorithms, the Weighted Guided Image Filter is used as a surround function instead of the Gaussian filter to estimate the background illumination, which can overcome the drawbacks of local blur and halo artifact that may appear by Gaussian filter. To avoid color distortion, the image is converted to the HSI color model, and only the intensity channel is enhanced. Then a linear color restoration algorithm is adopted to convert the enhanced intensity image back to the RGB color model, which ensures the hue is constant and undistorted. Experimental results show that the proposed method is effective to enhance both color and gray images with low exposure and non-uniform illumination, resulting in better visual quality than traditional method. At the same time, the objective evaluation indicators are also superior to the conventional methods. In addition, the efficiency of the proposed method is also improved thanks to the linear color restoration algorithm.

**Key words:** color image enhancement; weighted image guided filter; Retinex theory; linear color restoration algorithm

## 1 INTRODUCTION

Color images contain richer information than gray images. In recent years, color images have been applied more and more in many fields, such as intelligent traffic analysis, visual surveillance, and consumer electronics[1]. However, in practice, images are often obtained under undesirable weather and illumination conditions. Images taken under insufficient or non-uniform light shows low brightness, poor contrast, blurred local details, poor color fidelity and sudden changes in light, even often accompanied by a lot of noise. These make it difficult to extract and analyze information from images.

To enhance the color image of low illumination, it is required to maintain the color information without distortion while increasing the brightness and contrast, and highlight the image details and texture, so that the enhanced image is bright and natural. Conventional image enhancement methods are mostly aimed at gray images such as histogram equalization and its improved algorithms[2][3][4] in spatial domain enhancement, and wavelet transform[5][6][7] in frequency domain enhancement. A better result couldn't be achieved if these gray image enhancement algorithms are applied directly to color image due to a strong correlation between the RGB color channels. If each color channel is directly processed using the gray image enhancement algorithm, the different channels will be enhanced imbalancedly, which will lead to color distortion, saturation decrease and other problems.

In recent years, Retinex theory[8][9] based on color constancy has been widely used in image enhancement. In the conventional Retinex algorithm, the Single-Scale Retinex (SSR) or Multi-Scale Retinex (MSR) algorithm is used to process each RGB color channel separately. As mentioned above, the processed image will be subject to unexpected color distortion. Therefore, a Multi-Scale Retinex Algorithm with Color Restoration (MSRCR) is proposed by Jobson, et al[10], introducing a color restoration factor to correct the color distortion. Although MSRCR has achieved some effects in color rendition, it still does not completely solve the problem of color distortion.

To overcome the color distort problem, some scholars process images in other color models such as HSI, HSV, YCbCr, YUV, etc. In these color models, the brightness and color of the image are recorded in different independent channels. The processing of brightness channel will not affect color channel, ensuring no color shift occurs[11]. This method requires twice color model conversion, which takes more time. For example, Shin et al.[12][13] proposed an image enhancement method in HSV model. First, in the Value channel the illumination component is obtained by the Gaussian filter, and the reflection component is got by Retinex theory. Then the brightness of the illumination component is increased, and the processed illumination component is combined with reflection component again. Finally, the enhanced image is converted to RGB image. This method can better overcome the deficiencies such as color distortion and over enhancement. However, due to the isotropic characteristics of Gaussian filter when it is used to estimate the background illumination, the blurred edges of the resulting reflection image tend to be appeared, and the enhanced image is subject to halo artifact and low contrast. Thus, some researchers have tried to estimate the background illumination using a filter with anisotropy characteristics[14][15].

We have proposed a color image enhancement method based on Weighted Guided Image Filtering(WGIF). First, the image is converted to HSI color model, and WGIF is used to separate the illumination and reflection components; Second, the brightness and contrast of the illumination component is increased, and the reflection image is smoothed using WGIF; Then the processed illumination component and reflection component are fused into a new intensity channel image. Finally, a linear color restoration algorithm is adopted to convert the image back to the RGB color model. Due to the anisotropic properties and adaptive regularization term, the WGIF effectively avoids the local blur and halo artifact in the conventional Retinex algorithm. At the same time, the linear color restoration algorithm not only precisely preserves the color information of the original image, but also increases the computation efficiency.

## 2 RELATED WORK

Human perception can construct a visual representation with vivid color and detail across the wide dynamic range regardless lighting variations, it is color constancy. Illumination-reflection model believes the human visual perception of the color depends on the reflection characteristics of the object's surface, and the image can be mathematically represented as the product of the illumination components and the reflection component:

$$S_c(x,y) = L_c(x,y) \cdot R_c(x,y) \tag{1}$$

Where $c$ is one of the RGB color channels, that is $c \in \{R,G,B\}$. S, L, and R are the original image, the illumination image, and the reflection image, respectively.

Retinex theory is a model of the lightness and color perception of human vision. The main ideas of it is to calculate and eliminate the illumination component, only the reflection component is retained. However, the calculation of illumination component needs to solve an under-determined equation, which can not be accurately calculated and only can be estimated approximately. Through the years, the Retinex algorithm is evolved from path-comparison algorithms[16], iterative operation algorithms[17] to the center/surround algorithms[18]. Among them, the first two kinds of algorithms suffer from difficulty in adjusting parameters, high complexity and poor real-time performance. According to the central/surround Retinex algorithms, the color change caused by the illumination variation is smooth and belongs to the low frequency of the image; whereas the color change caused by the reflectivity is sharp and belongs to the high frequency part of the image. Therefore, the illumination component can be estimated by low-pass filtering.

Based on above analysis, Jobson et al.[19] put forward the Single-Scale Retinex (SSR) algorithm, in which the Gaussian Filtering(GF) is used as the center/surround function to estimate the background illumination, and a relatively accurate result has been obtained.

$$L_c(x,y) = S_c(x,y) * F_c(x,y) \tag{2}$$

$$F_c(x,y) = \frac{1}{2\pi\sigma} \exp[-\frac{(x^2+y^2)}{2\sigma^2}] \tag{3}$$

Where, $*$ denotes the convolution operation, and $F(x,y)$ is the Gaussian surround function, $\sigma$ is the scale parameter of function $F(x,y)$.

Take formula (2) into formula (1), and make logarithmic transformation to get the reflection component:

$$\log(R(x,y)) = \log(S(x,y)) - \log(S(x,y) * F(x,y)) \tag{4}$$

Exponential operation on $\log(R(x,y))$, the reflection component $R(x,y)$ can be obtained.

The scale parameter $\sigma$ in formula (3) is the only input parameter of the SSR algorithm, which directly impacts on the estimation result of the illuminance component. And SSR can either provide dynamic range compression(small scale $\sigma$), or tonal rendition(large scale $\sigma$), but not both simultaneously. Subsequently, the Multi-Scale Retinex algorithm (MSR) was proposed[10], which combines the dynamic range compression of the small-scale retinex and the tonal rendition of the large scale retinex. The MSR output is a weighted sum of the outputs of several different SSR output. Mathematically,

$$R_M(x,y) = \sum_{n=1}^{N} w_n \{\log(S(x,y)) - \log(S(x,y) * F_n(x,y))\} \tag{5}$$

where $N$ is the total number of scales; $w_n$ is the weight associated with the $n-th$ scale, and it needs meet the condition $\sum_{n-1}^{N} w_n = 1$.

The MSR is approaching human vision's performance in dynamic range compression but not quite achieve it. It fails to handle the images with regional or global graying-world violations effectively. In some cases, the "graying out" effect is severe and an unexpected color distortion occurs. Therefore, the Multi-Scale Retinex algorithm with Color Restoration (MSRCR)[10] was proposed, which provides good color rendition for images that contain gray-world violations. The mathematical expression is as follows:

$$R_{MSRCR_c}(x,y) = C_c(x,y)R_{MSR_c}(x,y) \quad (6)$$

$$C_c(x,y) = \beta\{\log[\alpha S_c(x,y) / \sum_{i=1}^{k} S_i(x,y)]\} \quad (7)$$

Where $C_c(x,y)$ is the color restoration function in the $c-th$ channel, which is used to adjust the color proportion of three color channels in RGB color model; $k$ refers to the number of RGB color channels with a value of 3; $\alpha$ controls the strength of the nonlinearity; $\beta$ is a gain constant. The MSRCR combines dynamic range compression of the small-scale Retinex and the tonal rendition of the large scale Retinex with a universally applied color restoration, which provides the necessary color restoration to a certain extent, eliminating the color distortions and gray zones evident in the MSR output.

## 3 COLOR IMAGE ENHANCEMENT METHOD BASED ON WGIF

The color image enhancement method proposed in this paper is as follows: First, the intensity channel ($S_I$) of HSI color model is obtained by the color-gray transform algorithm. Then the proposed enhancement method is applied in the intensity channel. Finally, the enhanced image is restored to the RGB color model by the linear color restoration algorithm.

## 3.1 Illumination Estimation

First, the intensity image $S_I$ is calculated by the mean-value method.

$$S_I = \frac{1}{3}(R(x,y) + G(x,y) + B(x,y)) \quad (8)$$

Where R, G and B denote the value of RGB channels respectively.

In the Retinex algorithm, the lighting is usually considered to be uniform, so the GF is used as center/surround function to estimate the illumination component. However, there may be abrupt lighting changes at the edges of the image[20], thus the gradient varies in all directions around the pixels are different. If the isotropic GF is used to estimate the illumination component, inaccurate results will be obtained in the light mutation region, resulting in the effect of halo artifact. Therefore, we need to preserve the edges of illumination while smoothing the other small fluctuations which is unrelated to that ones. Many studies have tried to estimate the illumination component by low-pass filters with anisotropic properties.

For example, some studies estimate illumination image by Bilateral Filtering(BF) and its improved algorithm[21][22], which can overcome some halo effect. But there are also some limitations in this method. On the one hand, the BF may suffer form "gradient reversal" artifacts in image enhancement[23][24][25]. The results may exhibit unwanted profiles around edges. That is because when a pixel on an edge has few similar pixels around it, the Gaussian weighted average is unstable. On the other hand, the efficiency of BF is poor. A brute-force implementation is $O(Nr^2)$ time with kernel radius $r$.

The Guided Image Filtering(GIF)[26] is derived from a local linear model, which computes the filtering output by considering the content of a guidance image. Therefore, it has the characteristic of anisotropy. The model assumes that the filtering output $q$ is locally a linear transform of the guidance image $I$ in a window $w_k$ centered at the pixel $k$:

$$q_i = a_k I_i + b_k, \forall i \in w_k \quad (9)$$

Where $i$ is the pixel index. $(a_k, b_k)$ are some linear coefficient assumed to be constant in $w_k$. Obviously, $\nabla q = a \nabla I$, in other words, the output image $q$ has an edge only if $I$ has an edge, and they have the same gradient direction. Therefore, the GIF avoids the gradient reversal artifacts that may appear in detail enhancement[27]. At the mean time, a main advantage of the guided filter over the BF is that it naturally has an O(N) time non-approximate algorithm for an image with $N$ pixels, independent of the window radius $r$ and the intensity range.

To determine the linear coefficient $(a_k, b_k)$ in formula (9), the minimum cost function model of input image $p$ and output image $q$ is established:

$$E(a_k, b_k) = \sum_{i \in w_k} ((a_k I_i + b_k - p_i)^2 + \varepsilon a_k^2) \tag{10}$$

Where $\varepsilon$ is a regularization factor penalizing large $a_k$, and it determines the criterion of a "flat path" or a "high variance" area.

The formula (10) is solved by the least square method, and it's solution is given by:

$$a_k = \frac{\frac{1}{|w|} \sum_{i \in w_k} I_i p_i - \mu_k \overline{p}_k}{\sigma_k^2 + \varepsilon} \tag{11}$$

$$b_k = \overline{p}_k - a_k \mu_k \tag{12}$$

Here, $\mu_k$ and $\sigma_k^2$ are the mean and variance of $I$ in $w_k$, $|w|$ is the number of pixels in $w_k$, and $\overline{p}_k$ is the mean of $p$ in $w_k$.

In the "High variance" region, the value of $a_k$ is larger, and the smaller $\varepsilon$ is required to penalize $a_k$; In the "Flat patch", the larger $\varepsilon$ is required to get smaller approximation error. Unfortunately, the value of $\varepsilon$ is fixed in the GIF, the blurring edge will still appear sometimes. In fact, halos are unavoidable for local filters[27].

The Weighted Guided Image Filtering(WGIF)[28] combines the advantages of global filtering and local filtering, which can adaptively adjust the regularization term based on the variance value in the local window. An edge weighting $\Gamma_I(i)$ is defined as follows:

$$\Gamma_I(i) = \frac{1}{N} \sum_{i'=1}^{N} \frac{\sigma_{I,1}^2(i) + \varsigma}{\sigma_{I,1}^2(i') + \varsigma} \tag{13}$$

Here, $I$ is a guidance image and $\sigma_I^2(\cdot)$ is the variance of $I$ in the $3 \times 3$ window. $N$ is the number of pixels; $\varsigma$ is a constant and its value is selected as $(0.001 \times L)^2$ while $L$ is the dynamic range of input image; $i'$ takes all pixels of the image.

The cost loss function described in formula (10) is transformed into:

$$E(a_k, b_k) = \sum_{i \in w_k} ((a_k I_i + b_k - p_i)^2 + \frac{\varepsilon}{\Gamma_I(i)} a_k^2) \tag{14}$$

For easy analysis, if the pixel $i$ is at an edge, the value of $\Gamma_I(i)$ is usually much larger than 1, thus the smaller $\frac{\varepsilon}{\Gamma_I(i)}$ is obtained in the formula (14), which can better preserve the edge of the image; if the pixel $i$ is at the flat area, the edge weight factor $\Gamma_I(i)$ is smaller, so a larger $\frac{\varepsilon}{\Gamma_I(i)}$ is obtained, and the smoothing effect is more obvious.

Thanks to the adaptive adjustment of regularization terms, WGIF has better behaviors near edges. it can preserve the edge details and avoid the halo artifact. Therefore, the reflection component can be calculated more accurately by WGIF. Besides, the computational complexity of the WGIF is $O(N)$, similar to that of GIF.

For these reasons, the WGIF is applied to the estimation of the illumination component in this paper. The intensity image of the original image is expressed as $S_I$, which will be used as the guidance image and the input image of the WGIF; and the output $\hat{q}$ of WGIF is the estimated illumination image, which is marked as $S_{IL}$.

$$S_{ILi}(x,y) = \hat{q}_i(x,y) = \overline{a}_i S_{I_i} + \overline{b}_i, \forall i \in w_k \quad (15)$$

To simplify the operation, the formula (1) is usually transformed into the logarithmic domain, and the reflection component is calculated by the formula (4). However, it may cause the gray information of the original image to be lost. Therefore, we directly compute the reflection component $S_{IR}$ from the estimated illumination component $S_{IL}$ based on the formula (1).

$$S_{IR}(x,y) = \frac{S_I(x,y)}{S_{IL}(x,y)} \quad (16)$$

The results of the illumination estimation experiment are shown in Fig.1. and Fig.2.

Fig.1 illustrates some examples of the estimated illumination by different filters. It can be observed that the result of GF is blurring at the step edge, which may lead some halo artifacts. The edge preserving effects of the BF and GIF are better than GF, but many details that are independent of the illumination are also preserved. In the result of WGIF, the strong edges are well preserved while the weaker textures are smoothed. Hence, a more accurate illumination component can be obtained by the WGIF compared to those above method.

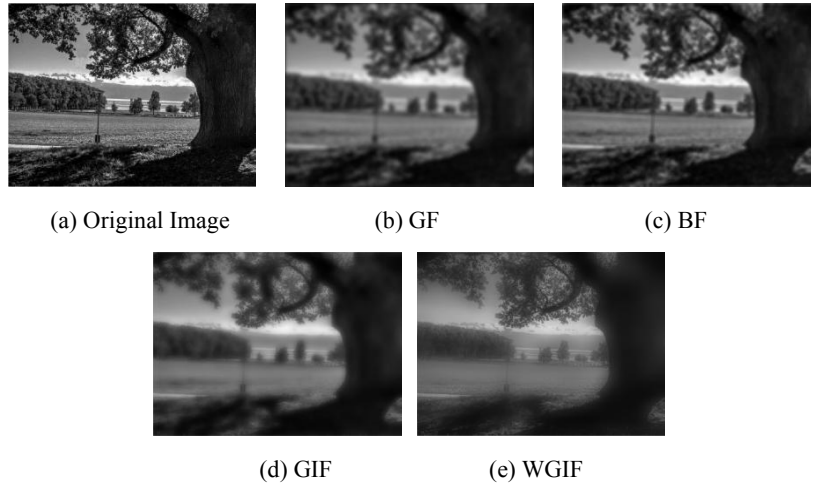

(a) Original Image    (b) GF    (c) BF

(d) GIF    (e) WGIF

**Fig.1. Comparison of the estimated Illumination by different filters**

Fig.2 shows the illumination and reflection components of HSI intensity channel. Fig2(a) is the intensity channel image. Fig2(b) and Fig2(c) is the result of GIF and WGIF. Let the intensity image be $S_I$, the illumination and reflection component be $S_{IL}$ and $S_{IR}$, respectively. It can be observed that illumination image estimated by GIF preserves more detailed textures, thus there are few details in the reflection image. By comparison, the illumination image estimated by WGIF is clear at the step edge, while the more abundant details are preserved in the reflection image.

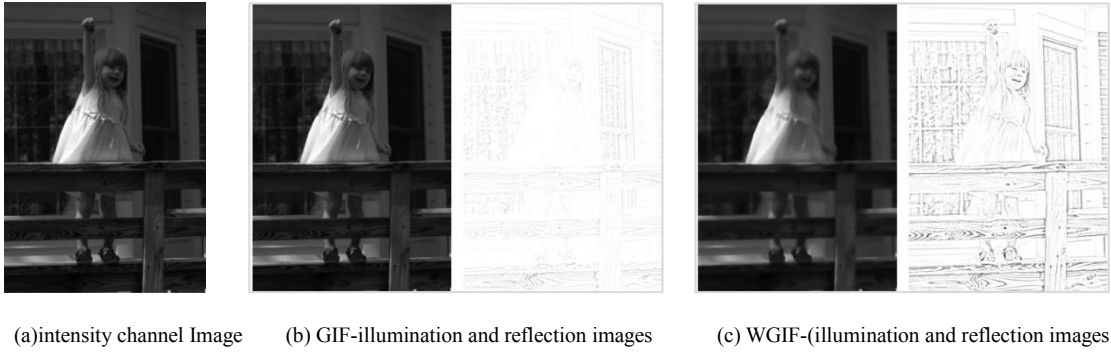

(a)intensity channel Image　　(b) GIF-illumination and reflection images　　(c) WGIF-(illumination and reflection images

Fig.2　Illumination and Reflection images of the HSI Intensity Channel

## 3.2 Adaptive Brightness Equalization

To avoid the loss of gray information, the proposed method dose not directly removed the illumination component in the logarithmic domain, but correct the illumination image and then combines it with the reflectance image. The brightness of the illumination image is very low. So it is corrected by the adaptive Gamma function[29] to improve the brightness of the image. the corrected illumination image is expressed as $S_{ILG}$ :

$$S_{ILG}(x,y) = (S_{IL}(x,y))^{\phi(x,y)} \qquad (17)$$

$$\phi(x,y) = \frac{S_{IL}(x,y)+a}{1+a} \qquad (18)$$

$$a = 1 - \frac{1}{m*n}\sum_{x=1}^{m}\sum_{y=1}^{n}S_{IL}(x,y) \qquad (19)$$

Where $m$、$n$ are the height and width of the original image respectively, $\varphi(x,y)$ is the Gamma correction function, and the parameter $a$ is the gray mean value of $S_{IL}$.

It can be known from the formula (17)-(19) that the algorithm can adaptively adjust the parameters of the Gamma correction by the value of illumination component. In the dark region the brightness is enhanced obviously, while the enhancement is suppressed in the bright region. The curve of Gamma correction function $y = x^r$ is shown in Fig.3. Where $r = (x+a)/(1+a)$ refers to the parameter value of the Gamma function, $a = 1 - \sum_{}^{n} x_i$ is the mean of input $x$.

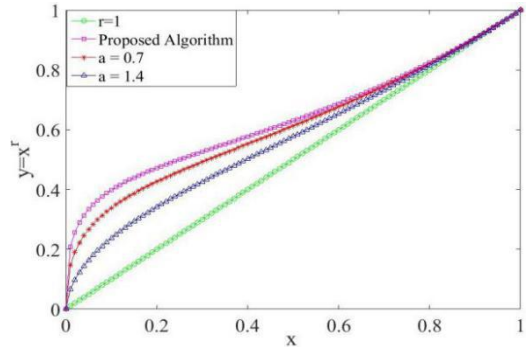

Fig.3. Adaptive Gamma Correction Curve

In Fig.3, it can be clearly seen that the adaptive Gamma correction algorithm can effectively enhance the intensity values of the dark areas. Therefore, it can exhibit more details. Howerver, the enhancement of bright areas is suppressed so that the details of the bright areas are preserved and the over-enhancement may be overcome

After adaptive Gamma correction, the contrast of the image is reduced. Thus a global linearly stretched is implemented on $S_{ILG}$, and the result image is $S_{ILGf}$.

## 3.3 Image Fusion

The noise corresponded to the higher frequencies in the image, so it mainly exists in the reflection component. To avoid being amplified, the noise of the reflection component need to be removed before image fusion. In this paper, $S_{IR}$ in the formula (16) is processed by the WGIF and the result is the denoised reflection image $S_{IRh}$:

$$S_{IRh} = q_{S_{IR_i}} = \overline{a}_i(S_{IR}) + \overline{b}_i \tag{20}$$

Then, the processed illumination image $S_{ILGf}$ is multiplied with the denoised reflection image $S_{IRh}$ to get the fused intensity image $S_{IE}$.

$$S_{IE}(x,y) = S_{ILGf}(x,y) \times S_{IRh}(x,y) \tag{21}$$

Finally, the S-hyperbolic tangent function[30] is used to improve the brightness of the fused image $S_{IE}$.

$$S_{IEf}(x,y) = \frac{1}{1+e^{-8*(S_{IE}-b)}} \tag{22}$$

$$b = \frac{1}{m*n}\sum_{x=1}^{m}\sum_{y=1}^{n}S_{IE}(x,y) \tag{23}$$

Here, $S_{IEf}$ is the enhanced intensity image; $m,n$ are high and wide of $S_{IE}$, respectively, and $b$ is the mean intensity value of $S_{IE}$.

## 3.4 Color Restoration

After the above operation, the enhanced intensity image is obtained. Now it needs to be re-converted to RGB color model. If the increase of each channel is inconsistent, it may lead to color distortion[31][32]. Therefore, the brightness gain coefficient[33][34] is calculated based on the original and enhanced intensity images, and then which is applied to image color restoration process. The brightness gain coefficient $\alpha$ is calculated as follows[35][36]:

$$\alpha(x,y) = \frac{S_{IEf}}{S_I} \tag{24}$$

The enhanced color image be convert back to the RGB color space based on $\alpha$ which ensures the linear proportions of the RGB channels in the original and enhanced color images remain unchanged. Moreover, the time cost of color space conversion is also reduced. The color linear restoration process is shown in the formula (25).

$$\begin{cases} R_1(x,y) = \alpha(x,y)R_0(x,y) \\ G_1(x,y) = \alpha(x,y)G_0(x,y) \\ B_1(x,y) = \alpha(x,y)B_0(x,y) \end{cases} \tag{25}$$

Here, it is assumed that the RGB three-channels of original and enhanced color images are

$[R_0, G_0, B_0]$, $[R_1, G_1, B_1]$ respectively.

## 4 ANALYSIS OF EXPERIMENTAL RESULTS

In order to verify the effectiveness of the proposed method, we compare the low illumination color images and gray images in different scenes. We use MATLAB 2015b for programming, and use a computer with an eight-core CPU, Intel 3.6GHz, 8G RAM, running on Windows 7. The following images are taken as examples to illustrate and compare the experiments. Three images (Tower, Apartment and Girl) from the NASA research center network and an image from the network (Mine) are used as the experimental objects for the enhancement of color images. CMU-PIE and YaleB face databases are used as the verification object of gray image enhancement algorithm, where Face1 and Face3 came from CMU-PIE database, Face2 and Face4 came from YaleB database. Since we only care about the illumination problem, in the YaleB face database, we only select the same face in same pose and different light. (pose is P00, light is [0°~77°]). The color enhancement algorithms for comparison include SSR, MSR, MSRCR and the proposed method in this paper; and the gray image enhancement algorithms for comparison are SSR, MSR and the algorithm. The parameters for all experiments are set as follows: SSR—the scale parameter $\sigma = 80$; MSR and MSRCR—the scale parameter $\sigma = 15, 80, 250$, the number of Gaussian functions $N = 3$, the weighting coefficients $w_1 = w_2 = w_3 = \frac{1}{3}$; Proposed algorithm—the window radius $r = 5$, regularization factor $\varepsilon = 0.1^2$, constant $\varsigma = 0.065536$ in the WGIF.

## 4.1 Subjective Evaluation

Fig.4 and Fig.6 are the results of the low illumination color image enhancement both with uniform and non-uniform lighting. Fig.5 is the zoom-in patch of the Tower image in Fig.4. It can be seen, the brightness enhancement effect of SSR algorithm is limited, while MSR algorithm over-enhances the image, as a result, some details in the brighter area are lost, the noise is more obvious, and the halo appears in the step edge obviously. (As shown in the MSR result of the Tower image in Fig.5). Compared with SSR and MSR algorithms, MSRCR algorithm improves brightness and preserves the color. However, the contrast of the enhanced image is low and some details are lost. Moreover, MSRCR suffers from the most severe halo (As shown in the MSRCR result of the Tower image in Fig.5). The proposed algorithm improves the local brightness and contrast adaptively, the dark area is effectively enhanced, while the details of bright area remain well. In addition, the halo problem is also avoid, and the color of the enhanced image is vivid and natural.

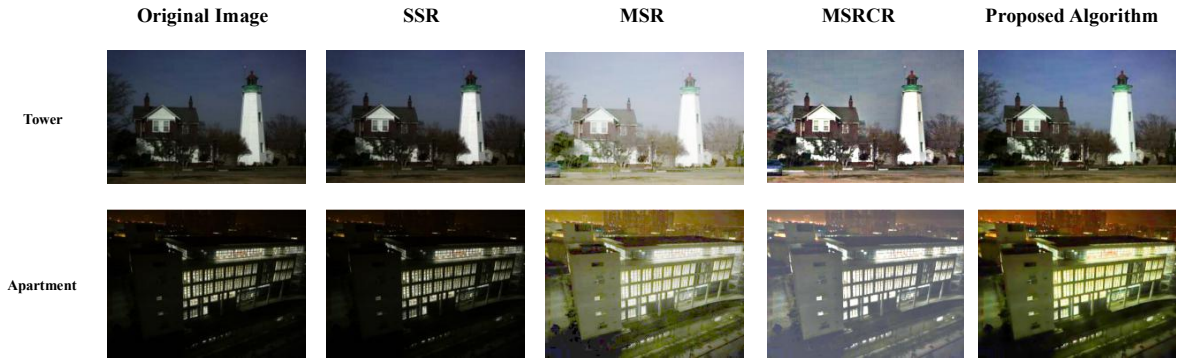

**Fig.4 Low illumination color image enhancement under with uniform light**

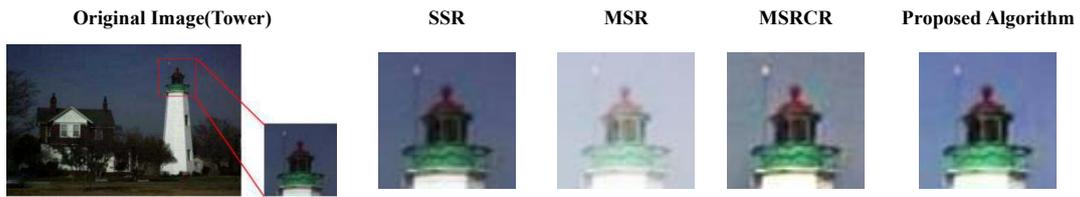

Fig.5 Detail contrast of Tower image enhancement

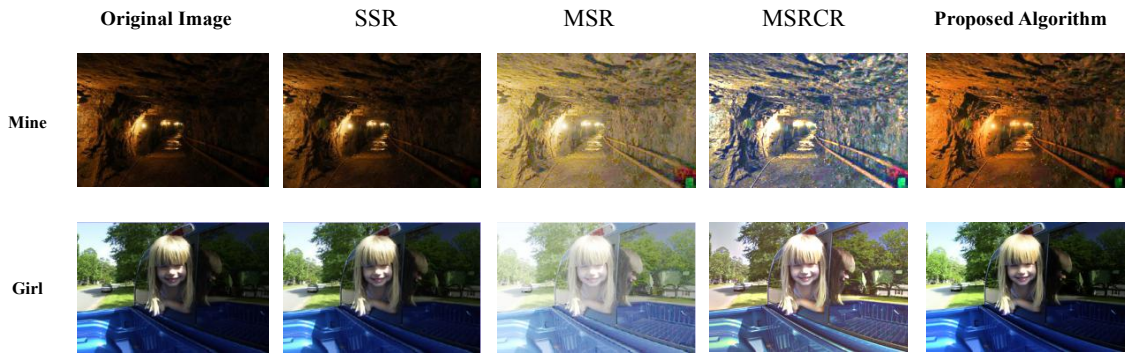

Fig.6 Low illumination color image enhancement with non-uniform light

It can be seen, SSR algorithm improves the brightness of the images under uniform illumination, but it doesn't work effectively of the images under non-uniformly illumination, and the noise is also amplified. The MSR algorithm over-enhanced the iamge so that the contrast is decreased, and some details are lost. It can be observed, the noticeable halo artifact appears in the sharp edge (as shown in Fig.9), and the noise is very obvious. By comparison,the proposed algorithm improves both brightness and contrast, so that the image is bright and clear. But for images under very non-uniform lighting, the enhancement effect in dark area is still limited.

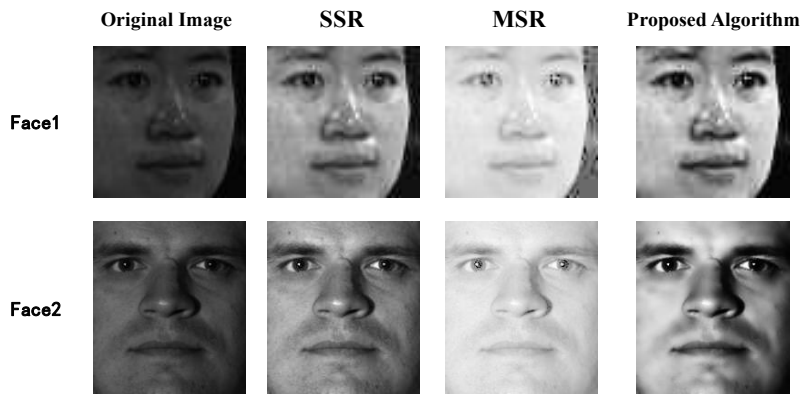

Fig.7 Low illumination gray image enhancement with uniform light

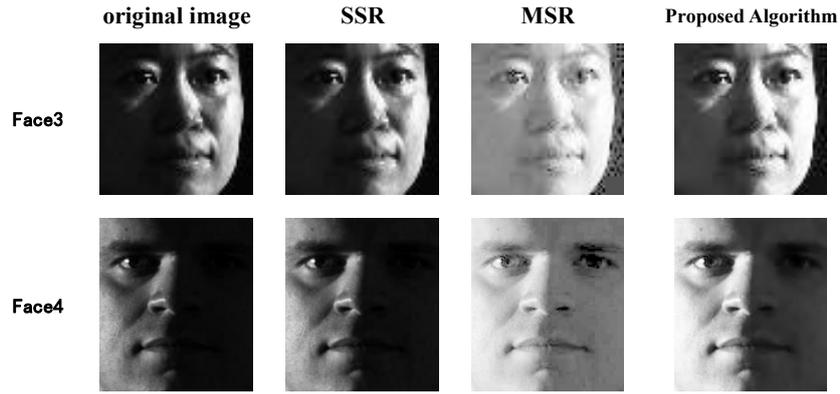

Fig.8 Low illumination grayscale image enhancement with non-uniform light

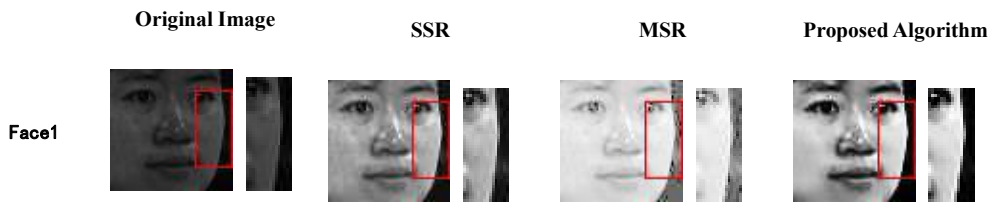

Fig.9 Enhancement detail contrast of Face1

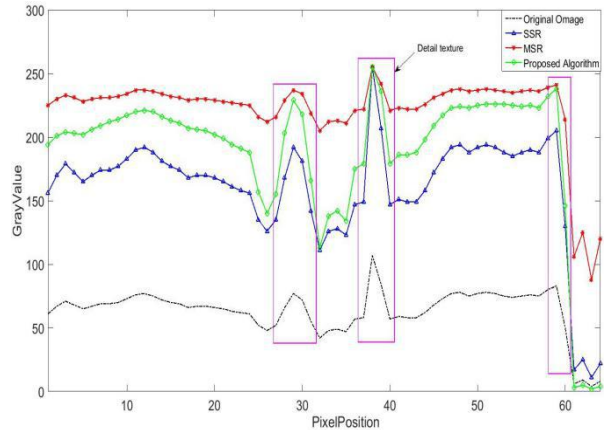

Fig.10 Pixel scan results of the 30th line of image Face1

Fig.10 is the scan result of the 30th line pixel of the image Face1 processed by various algorithms. It can be seen that the pixel intensity values in the original image are lower, and their values are improved after being processed by SSR, MSR and the proposed algorithm. Among them, the MSR algorithm maximally improves the brightness value, but the curve is the most flat, that is to say, the contrast is the lowest. The SSR algorithm slightly improves the intensity value, but the smoothing effect on the low frequency region is poor so that the noise is obvious. The proposed algorithm enhances the brightness values moderately, The intensity values are significant increased in the high frequency region, and are well smoothed in the low frequency region.Thus it can display image details in the dark area and the noise is removed.

## 4.3 Objective Evaluation

In this part, some objective indexes such as information entropy, brightness, contrast, mean gradient, edge intensity, and the $std \times gray$ [37] are adopted to evaluate the different enhancement algorithm. To calculate $std \times gray$, the image is divided into non-overlapping blocks with the same size. The standard deviations of all the blocks are averaged to get the average variance $std$. Then multiply the $std$ multiply and the mean value $gray$ of all the blocks. The larger the value of the $std \times gray$, the higher the quality of the image. When evaluating color image, the indexes △B[36], △C[36] and △H[32] is used to measure the change rates of the brightness, contrast, and hue. The calculations are as follows:

$$\Delta B = \frac{Mean(I_{out}) - Mean(I_{in})}{Mean(I_{in})} \quad (26)$$

$$\Delta C = \frac{Var(I_{out}) - Var(I_{in})}{Var(I_{in})} \quad (27)$$

$$\Delta H = abs\left(\frac{Mean(H_{out}) - Mean(H_{in})}{Mean(H_{in})}\right) \quad (28)$$

Where $I_{in}, I_{out}$ refer to the intensity channel of the original and enhanced color images in the HSI color model. $H_{in}$, $H_{out}$ are the hue channel of the original and enhanced images. $\Delta H$ describes the change rate of the color. Obviously, lager $\Delta B$, $\Delta C$, and smaller $\Delta H$ is expected.

The objective evaluation results of the color and gray images in low illumination environment are shown in Table.1 and Table.3, respectively. It can be obtained seen, the results of information entropy, $\Delta C$, average gradient, and edge intensity of the proposed algorithm are obviously superior to other algorithms, and the $\Delta H$ in color image is much less than that of other algorithms, which shows that the proposed algorithm best preserves the color information. Though the $\Delta B$ of proposed algorithm is lower than that of MSR or MSRCR, combining with the visual effect of the image, we know that just increase brightness does not mean good enhancement, which will lead to over-enhancement, resulting in loss of image details.

Table.2 compares the efficiency of the the color mode conversion algorithm. Clearly, the efficiency of linear color restoration algorithm is higher than that of non-linear color restoration algorithm.

**Table.1. Objective Quality Evaluation of Different Enhancement Algorithms for Low Illumination Color Images**

| | | Information Entropy | △B | △C | △H | Average Gradient | Edge Intensity | $Std \times Gray(\times 10^4)$ |
|---|---|---|---|---|---|---|---|---|
| mine/ girl | Original-image | 5.85/6.64 | --- | --- | --- | 3.47/3.69 | 30.374/32.48 | 1.970/9.72 |
| | SSR | 6.22/6.77 | 0.31/0.17 | 0.32/0.34 | 0.032/0.004 | 4.31/4.07 | 37.44/35.73 | 3.347/16.99 |
| | MSR | 7.09/6.89 | 4.33/*1.54* | 1.01/-0.09 | 0.325/0.035 | 8.12/4.86 | 68.49/41.09 | 16.55/24.09 |
| | MSRCR | 7.27/7.13 | *4.59*/0.92 | *2.32*/0.61 | 3.417/0.068 | 7.35/6.56 | 68.86/58.63 | 17.59/31.84 |
| | Proposed-algorithm | *7.41/7.36* | 1.88/0.29 | 0.92/*0.72* | 7.61*10^-5/2.02*10^-4 | *14.92/8.54* | *124.15/74.72* | *29.22/34.20* |
| tower / Apartment | Original-image | 7.14/5.44 | --- | --- | --- | 8.82/8.04 | 67.01/63.78 | 30.77/3.69 |
| | SSR | 7.36/5.78 | 0.34/0.24 | 0.04/0.26 | 0.001/0.006 | 9.51/8.92 | 71.83/63.82 | 37.41/5.605 |
| | MSR | 7.41/6.93 | *2.09*/3.80 | -0.51/-0.79 | 0.002/0.026 | 9.17/11.95 | 66.06/89.94 | 39.33/28.56 |
| | MSRCR | 7.45/6.27 | 1.18/*4.94* | -0.36/0.11 | 0.023/1.414 | 10.83/10.49 | 77.37/83.242 | 38.96/22.13 |
| | Proposed-algorithm | *7.49/7.45* | 0.81/2.45 | *0.10/1.30* | 6.95*10^-5/2.08*10^-5 | *14.07/12.33* | *111.79/98.142* | *45.65/32.75* |

Table.2. Efficiency of color model conversion algorithm(Unit:s)

| Method \ Image Effectiveness | Mine | Girl | Tower | Apartment |
|---|---|---|---|---|
| Non-linear Color Restoration | 0.356601 | 0.128500 | 0.120166 | 0.089173 |
| Linear Color Restoration | 0.167030 | 0.101748 | 0.092218 | 0.079519 |

Table.3. Objective quality evaluation of different enhancement algorithms for low illumination gray images

| | | Information Entropy | Brightness | Contract | Average Gradient | Edge Intensity | $Std \times Gray(\times 10^4)$ |
|---|---|---|---|---|---|---|---|
| Face1/ Face2 | Original-image | 6.13 / 6.64 | 48.94/65.59 | 20.36/25.58 | 5.24/2.97 | 44.71/27.78 | 2.03/4.29 |
| | SSR | 6.13 /6.64 | 126.09/117.55 | 50.10/42.08 | 12.72/4.88 | 108.32/45.70 | 31.65/20.82 |
| | MSR | 5.99/6.18 | **204.16/205.43** | 38.49/25.04 | 9.22/3.07 | 71.65/27.74 | 30.24/12.88 |
| | Proposed-algorithm | 7.48 / *7.78* | 143.91/128.97 | 65.16/63.77 | 15.94/5.22 | 139.81/54.04 | 70.33/52.45 |
| Face3/ Face4 | Original-image | 6.79/6.42 | 64.06/46.69 | 71.73/46.81 | 14.20/2.76 | 118.31/26.11 | 39.26/ 10.23 |
| | SSR | 6.72/6.42 | 74.31/65.00 | 74.75/60.69 | 14.79/3.51 | 122.64/33.29 | 41.52/23.94 |
| | MSR | 6.28/6.08 | *163.32/156.26* | 53.13/53.59 | 13.27/3.58 | 96.32/33.50 | 46.09 / 44.87 |
| | Proposed-algorithm | *7.16/7.33* | 100.20/116.85 | *90.45/82.17* | *16.32/4.75* | *137.18/46.47* | *81.96/78.90* |

## 5 CONCLUSION

A novel low illumination image enhancement method based on Retinex is presented in this paper. WGIF is used to estimate illumination and remove noise, which effectively overcome some problems such as halo artifact, detail loss and noise amplification. To avoid color distortion, the proposed algorithm processed the image in the intensive channel in HSI color model and the linear color restoration is used. The linear color restoration algorithm not only ensures the hue is constant and undistorted but also helps to achieve higher operation efficiency. To avoid the loss of gray level information, the algorithm does not calculate and eliminate the illumination component in the logarithmic domain. Instead, it enhances the brightness adaptively according to the illumination image, which effectively avoids the over enhancement of the bright area. The experimental results show that the proposed algorithm can efficiently enhance both color and grey imges with low illuminate. Both subjective and objective evaluation have achieved good results. Nevertheless, When the illumination is very uneven, the enhancement of the local dark region is limited, and it requires further research.